\pdfoutput=1
\documentclass[11pt]{article}

\usepackage[final]{acl}

\usepackage{times}
\usepackage{latexsym}

\usepackage[T1]{fontenc}
\usepackage[utf8]{inputenc}

\usepackage{microtype}

\usepackage{inconsolata}

\usepackage{graphicx}

\usepackage{algorithm}
\usepackage{algpseudocode}
\usepackage{amsmath}
\usepackage{caption}
\usepackage{amssymb}
\usepackage{mathtools}
\usepackage{amsthm}
\usepackage{colortbl}
\usepackage{multirow}

\title{Aligning Large Language Models with Diverse Political Viewpoints}

\author{Dominik Stammbach \\
  Princeton University \& \\
  ETH Zurich \\
  dominsta@princeton.edu \\\And
  Philine Widmer \\
  Paris School of Economics \& \\ ETH Zurich \\ 
  \texttt{philine.widmer@psemail.eu} \\\And
  Eunjung Cho \\
  ETH Zurich \\
  \texttt{choeun@ethz.ch} \\ \AND
  Caglar Gulcehre \\
  EPFL Lausanne \\
  \texttt{caglar.gulcehre@epfl.ch} \\\And
  Elliott Ash \\
  ETH Zurich \\
  \texttt{ashe@ethz.ch} \\}

\begin{document}
\maketitle
\begin{abstract}

Large language models such as ChatGPT exhibit striking political biases. If users query them about political information, they often take a normative stance. To overcome this, we align LLMs with diverse political viewpoints from 100,000 comments written by candidates running for national parliament in Switzerland. Models aligned with this data can generate more accurate political viewpoints from Swiss parties, compared to commercial models such as ChatGPT. We also propose a procedure to generate balanced overviews summarizing multiple viewpoints using such models. The replication package contains all code and data.

\end{abstract}

\hspace{3mm}
\begin{minipage}[c]{5mm}
\includegraphics[width=\linewidth]{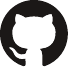}
\end{minipage}
\hspace{1.5mm}
\begin{minipage}[c]{0.7\textwidth}
\fontsize{0.76em}{0.76em}\selectfont
% Code: 
\begin{minipage}[c]{0.7\textwidth}
    \href
     {https://github.com/dominiksinsaarland/aligning-LLMs-with-political-views}
     {\texttt{github.com/dominiksinsaarland/aligning-}}
    
     \,\,\,\,\href
     {https://github.com/dominiksinsaarland/aligning-LLMs-with-political-views}
     {\texttt{LLMs-with-political-views}}
 \end{minipage}
\end{minipage}
\vspace{-3mm}

\section{Introduction}

Large language models (LLMs) have become very popular, with chat applications like ChatGPT and Gemini having hundreds of millions of active users combined.\footnote{\tiny{\url{https://www.demandsage.com/chatgpt-statistics/}\\\indent\indent \url{https://www.demandsage.com/google-gemini-statistics/}}}  One of the intended use cases is the retrieval of factual information \citep[e.g.,][]{mehdi2023}. Interacting with chatbots can influence users' views \cite{Jakesch2023} and potentially influence behavior \citep[e.g.,][]{stieger2021, sharma2024generativeechochambereffects}. Because of this, LLMs -- if used as decision aids in high-stakes contexts such as shaping political views or votes -- must return factually correct and unbiased statements.

Political bias is present in all first-generation LLMs \cite{feng-etal-2023-pretraining}. Also, ChatGPT is not impartial, as several recent papers have shown: ChatGPT exhibits progressive, liberal, and pro-environmental biases \cite{rozado2023political, hartmann2023political, Motoki2024, Rutinowski2024}. Given these findings, \citet{hartmann2023political} ask: "What if ChatGPT exhibits a political ideology that may pervade its synthetic responses and subtly influence its millions of unsuspecting users?"

\begin{figure}[t!]
    %\vspace{-1mm}\hspace*{-2.7mm}
    \centering
%    \small
    \begin{tabular}{p{7.5cm}} 
\textbf{system:} You are a helpful Swiss policy advisor. You are asked a policy issue or question. You are in the political party \textbf{P}, and you reply in \textbf{L}. \\
\textbf{user:} What's your opinion on the following issue or question: \textbf{Q} \\
    \end{tabular}
    \vspace{-2.5mm}
    \caption{Prompt for conditional generation. Varying attributes are party P, language L, and political issue Q.}% For example, party = "GLP", language = German, and political issue = Should the state do more to promote equal educational opportunities?.}
    \label{fig:example_prompt}
\end{figure}

To address such concerns, we propose to align LLMs such that they can generate more diverse political viewpoints. Inspired by early work with GPT models and conditional generation based on metadata \citep{NEURIPS2019_3e9f0fc9}, we align LLMs toward accurately reflecting diverse viewpoints from Swiss parties. Throughout the paper, we use prompts of the form shown in Figure \ref{fig:example_prompt}: We ask an LLM about its opinion on political issues while being in party \textit{P} and speaking language \textit{L}.\footnote{Switzerland has four official national languages.}

We align LLMs with data obtained from the Swiss voting advice application \textit{smartvote}: We have ca. 100,000 comments written by candidates running for national parliament in Switzerland, coupled with meta-data, comprising 33 attributes such as party, language, canton, age, profession, favorite movies, and others. Smartvote lets both user and candidates fill out a survey about stances toward various political issues. Then, it matches candidates with the highest overlap to a user. This helps voters decide which candidates they could vote for.

We took that survey with ChatGPT. We find that ChatGPT is most closely aligned with candidates from the \textit{Green Liberal Party of Switzerland}, followed by other liberal parties. These results confirm its progressive and pro-environmental bias in the Swiss case. A more exhaustive analysis is presented in Appendix \ref{sec:political_bias_chatgpt}, and we show anecdotal evidence of such biases in Table \ref{tab:example_generations}.

%\citep[which has been shown for other countries, e.g.,][] {rozado2023political, hartmann2023political, Motoki2024, Rutinowski2024} in the Swiss context. We present anecdotal evidence of such biases in Table \ref{tab:example_generations} 

\begin{table*}[t!]
    \centering
    \scriptsize
    \begin{tabular}{l p{9cm} p{3cm}}
         Party & ChatGPT generation & ORPO-aligned Llama 3 model  \\ \hline
         SP & Yes, the state should do more to promote equal educational opportunities in order to promote equal opportunities and reduce social inequalities. Tuition vouchers for pupils from low-income families can be a useful instrument in this regard. \textbf{overlap answer and party stance: high} & Equal opportunities are the foundation of our state. \textbf{overlap answer and party stance: high} \\
         Die Mitte & Yes, Die Mitte supports the promotion of equal educational opportunities for all pupils. Measures such as tuition vouchers for pupils from low-income families can help to promote equal opportunities in the education system. \textbf{overlap answer and party stance: medium}&  Funding is important, but not only for low-income families. \textbf{overlap answer and party stance: high}\\
         FDP & Yes, the state should do more to promote equal educational opportunities. Tuition vouchers for pupils from low-income families could be a useful instrument in this regard. \textbf{overlap answer and party stance: low} & The cantons are responsible for providing schools. \textbf{overlap answer and party stance: high}\\ 
         SVP & Yes, the state should do more to promote equal educational opportunities in order to promote equal opportunities and reduce social inequalities. Tuition vouchers for pupils from low-income families can be a useful instrument in this regard. \textbf{overlap answer and party stance: low} & Promoting low-achieving pupils is not the solution. \textbf{overlap answer and party stance: high}\\ \hline
         
    \end{tabular}
    \caption{Political stances generated with ChatGPT and an aligned model for the policy issue \textit{Should the state do more to promote equal educational opportunities?} for all major Swiss parties represented in the Federal Council of Switzerland. Political leanings (taken from Wikipedia): SP = center-left, Die Mitte = centerist, FDP = center-right, SVP = right-wing. Text in bold (overlap) inserted by authors.}
    \label{tab:example_generations}
\end{table*}

% expand on this for a bit
%To overcome such implicit biases present in LLMs, 
In this work, we align Llama 3 models \citep{llama3modelcard} with smartvote data, combining conditional generation \citep[e.g.,][]{NEURIPS2019_3e9f0fc9} and monolithic preference optimization \citep{hong2024orpo} alignment. We find that the resulting aligned models generate more diverse and more accurate political viewpoints compared to commercial or non-aligned models -- and these aligned viewpoints are preferred in human evaluation.

Such models can be used to create accurate political views of all Swiss parties toward an issue, which then could be summarized by other capable LLMs \citep[e.g.][]{openai2024gpt4} to give balanced overviews. Such approaches potentially facilitate finding political compromises or learning more about political issues. However, we urge more research to better understand the promises and dangers of AI in providing political information or voting advice. In any case, we strongly believe that if LLMs were used in such circumstances, they'd better be accurate and impartial.

\section{Related Work}

After LLMs became popular, it did not take long for social scientists to start investigating the political leaning of LLMs, specifically ChatGPT \citep[e.g.,][]{rozado2023political, hartmann2023political, Motoki2024, Rutinowski2024}. All of them reported that ChatGPT has political biases and a certain political leaning.

Relatedly, the NLP community also noticed political leanings of LLMs \citep{feng-etal-2023-pretraining, bang-etal-2024-measuring}, and started to develop appropriate counter-measures. One approach is to explicitly align them with specific leanings: \citet{jiang-etal-2022-communitylm} train "CommunityLM", an LM specifically aligned with a certain political leaning on a dedicated corpus -- and the authors investigate the worldviews of such communities by probing the resulting LLM. Concurrent work fine-tunes an ensemble of such CommunityLMs, and shows that these can be used as well to produce balanced overviews \citep{feng2024modularpluralismpluralisticalignment}. Lastly, future work to generate balanced overviews using our LLMs would profit from literature on summarizing subjective opinions \cite{suhara-etal-2020-opiniondigest, amplayo-lapata-2021-informative} or explicitly generating consensus statements \citep{consent_statements}.

\section{Data}

We use the same data source as \cite{vamvas2020xstance}: Comments written by candidates running for national parliament in Switzerland. The comments were written for and submitted to the voting advice application \textit{smartvote}. This application helps voters determine which candidates or parties have political preferences similar to their own. Prior to an election, candidates can report their stance on a short (30 questions) or long (75 questions) survey across various political issues. Voters can take the same survey and are matched with candidates having the highest overlap (smartvote returns an ordered list of all candidates a voter can elect, ordered by survey overlap). The questions are drafted by a team of political scientists \citep[for more details, see][]{thurman2009smartvote}. These questions are yes-no questions, and smartvote allows candidates to submit comments which further explain their stance on an issue (these responses can be queried by users). We have 100,000 such comments for ca. 200 political questions across the last three national elections in Switzerland. We use these comments and metadata to align models using conditional generation. 

Smartvote is a popular service in Switzerland: 85\% of candidates running for elections in Switzerland have a smartvote profile, and one in five voters consults smartvote before elections. Thus, our data likely offers a rich overview of possible political stances in Switzerland. We show more detailed dataset statistics in Appendix \ref{sec:dataset_stats}.

\section{Methods}

In conditional generation, we want to generate text based on constraints or metadata \citep[e.g.,][]{NEURIPS2019_3e9f0fc9, zhou2023controlled}. For example, previous work generated news articles based on the attributes \textit{domain, date, authors} and \textit{headline} \citep{NEURIPS2019_3e9f0fc9}. Alignment datasets usually contain triples of the form \textit{instruction, preferred choice} and \textit{rejected choice}.\footnote{That is for reference-free methods such as DPO \citep{tunstall2023zephyr} or ORPO \citep{hong2024orpo}.} We interpret conditional generation for alignment as sampling a comment toward a political issue \textit{q} drafted by somebody from party \textit{p} speaking language \textit{l} as the preferred choice. For the rejected choice, we sample a comment for the same issue \textit{q} in the same language \textit{l}, but from a candidate  who is in a different political party $\neg$\textit{p}.%

We use reference-free monolithic preference optimization \citep[ORPO;][]{hong2024orpo} as our alignment objective. We optimize the following joint loss   taken directly from the ORPO paper:
\begin{equation}
    \mathcal{L}_{ORPO} = \mathbb{E}_{(x, y_w, y_l)}\left[ \mathcal{L}_{SFT} + \lambda \cdot \mathcal{L}_{OR} \right]
    \label{eq:main}
\end{equation}

\begin{equation}
    \mathcal{L}_{OR} = -\log \sigma \left( \log \frac{\textbf{odds}_\theta(y_w|x)}{\textbf{odds}_\theta(y_l|x)} \right) \label{eq:ratio} 
\end{equation}

where the first part of equation \ref{eq:main} is just supervised fine-tuning. The second part $\mathcal{L}_{OR}$ from equation \ref{eq:ratio} increases the likelihood of the preferred choice $y_w$ and decreases the likelihood of the rejected choice $y_l$. For the exact details, we refer to \cite{hong2024orpo}. We believe this loss is well suited for conditional generation, as it pushes apart comments with different metadata, although they might only differ in subtle nuances. We will compare ORPO-aligned models to direct supervised fine-tuning \citep[dSFT;][]{taori2023alpaca} in the results section.

We also experimented with direct policy optimization DPO \cite{rafailov2023direct} following the recipe outlined in \citep{tunstall2023zephyr} and RLHF \citep{stiennon2022learning}. However, initial qualitative exploration has shown that the trained models did not generate satisfactory output. We use LORA for all experiments but do not tune hyperparameters. We believe that it should be possible to get similar results with other alignment algorithms, but we have not explored that any further.

In all our experiments, we used the transformer TRL library \cite{vonwerra2022trl} and the 4bit quantized unsloth version of Llama 3 8B models\footnote{\url{https://huggingface.co/unsloth/llama-3-8b-Instruct-bnb-4bit}} \citep{llama3modelcard} and fine-tuned all models using LoRA \citep{hu2022lora}. For the supervised fine-tuning, we used the hyper-parameters outlined in \citep{tunstall2023zephyr}, and for ORPO alignment, we proceeded with the hyper-parameters outlined in \citep{hong2024orpo}.

\section{Results}

We present four sets of results on our dataset's development and test split: Qualitative evidence, diversity of generations, similarity between generated text and human references, and human evaluation. All our models use the prompt template shown in Figure \ref{fig:example_prompt}.

We present results for zero-shot settings (ChatGPT 3.5, GPT-4o, and Llama 3), few-shot settings (ChatGPT 3.5), Llama 3 fine-tuned using direct supervision \citep{taori2023alpaca}, and Llama 3 aligned using ORPO.

\subsection{Diversity of Generations}

We show qualitative evidence of political bias in ChatGPT generations and a lack of variety in responses in Table \ref{tab:example_generations}, where ChatGPT generates almost identical, progressive responses for all parties, although actual party stances toward the issue of whether the state should promote equal educational opportunities vary substantially. The ORPO-aligned Llama 3 model, on the other hand, accurately captures these different stances.

%, and we present results for the following models:  (1) ChatGPT 3.5 zero-shot, (2) Llama-3-instruct zero-shot, (3) Llama-3-instruct-finetuned-with-dSFT \citep{taori2023alpaca} and (4) Llama-3-instruct-aligned-with-ORPO \citep{hong2024orpo}.

\begin{figure}[h!]
    \centering
    \includegraphics[width=0.9\linewidth]{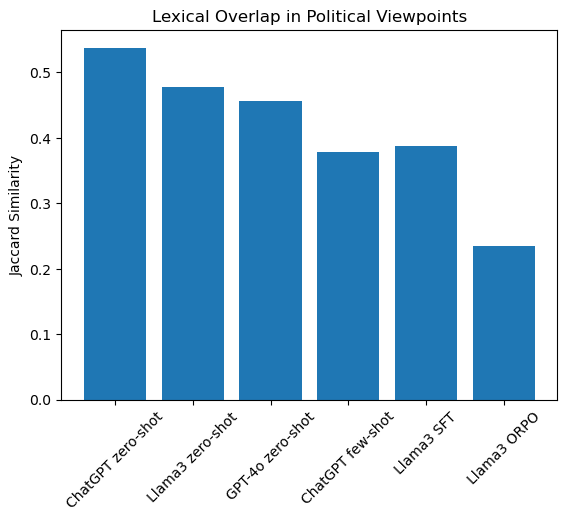}
    \caption{Average diversity of replies within a political issue, measured with Jaccard similarities (lower Jaccard similarity means higher diversity).}
    \label{fig:average_diversity}
\end{figure}

%We show qualitative evidence of political bias in ChatGPT generations and a lack of variety in responses in Table \ref{tab:example_generations}, where ChatGPT generates similar progressive responses for all parties, although party stances vary substantially. 
We present additional quantitative evidence of this phenomena. For each political issue and model, we compute Jaccard similarities between the generations for different parties. We plot the average overlap between responses measured in Jaccard similarities\footnote{Jaccard similarity computes the intersection divided by the union. If two sets contain the same words, this similarity is 1. If there exist no overlapping words, this similarity is 0.} in Figure \ref{fig:average_diversity}.

We find that all zero-shot generations are strikingly similar and have a high average similarity. Both few-shot learning and supervised fine-tuning results in more diverse generations, reducing the number of overlapping words by 30\% compared to ChatGPT zero-shot. The ORPO-aligned models further reduce overlapping generations and result in an average similarity of 0.24, roughly half of the overlap measured for ChatGPT 3.5 in a zero-shot setting.

\subsection{Quantitative Evaluation}

As another set of results, we compute MAUVE scores \citep{pillutla-etal:mauve:neurips2021}. MAUVE is an automated metric that measures the gap between neural text and human references using LLM representations. The higher the MAUVE score, the closer the generated text and the human references are. Because our generations are either in German, French, or Italian, we use a multi-lingual RoBERTa model as a featurizer \cite{conneau-etal-2020-unsupervised}. Table \ref{tab:mauve_results} shows the resulting scores over different dataset splits. We show average results over five runs (with 95\% confidence intervals), sampling different reference comments in each run.

\begin{table}[h!]
    \centering
    \tiny
    \begin{tabular}{l c c c}
    Model & MAUVE dev & MAUVE test & MAUVE (dev + test) \\ \hline
    ChatGPT zero-shot & 0.36 $\pm$ 0.02 & 0.25 $\pm$ 0.05 & 0.24 $\pm$ 0.02 \\
    Llama 3 zero-shot & 0.27 $\pm$ 0.05 & 0.03 $\pm$ 0.0 & 0.08 $\pm$ 0.01 \\
    GPT-4o zero-shot & 0.22 $\pm$ 0.02 & 0.25 $\pm$ 0.03 & 0.16 $\pm$ 0.01 \\
    ChatGPT few-shot & 0.49 $\pm$ 0.03 & 0.59 $\pm$ 0.02 & 0.49 $\pm$ 0.01 \\
    Llama 3 SFT & 0.48 $\pm$ 0.02 & 0.48 $\pm$ 0.03 & 0.38 $\pm$ 0.02 \\ 
    Llama 3 ORPO & \textbf{0.63} $\pm$ 0.03 & \textbf{0.71} $\pm$ 0.05 & \textbf{0.64} $\pm$ 0.01 \\ \hline
    \end{tabular}
    \caption{Automated metrics measuring overlap between model-generated replies and actual replies in the development and testset.}
    \label{tab:mauve_results}
\end{table}

The overall picture is similar to the diversity results. Zero-shot experiments result in the lowest overall MAUVE scores. Few-shot and supervised fine-tuning again lead to comparable MAUVE scores. Lastly, the ORPO-aligned generations obtain the by far highest MAUVE scores, indicating that they are closest to the actual reference comments. %Perhaps not surprisingly, the zero-shot answers are most distant from human references, arguably because zero-shot models do not know what such comments look like. Both SFT and  

These results are robust across runs, and the 95\% confidence intervals remain small. Furthermore, we computed MAUVE scores with an MBERT encoder \citep{devlin-etal-2019-bert} which produces very similar results and the same ranking \citep{devlin-etal-2019-bert}.

\subsection{Human Evaluation}

Finally, we perform human evaluation of the generated comments. Each annotated datapoint consists of an instruction (see Figure \ref{fig:example_prompt}) and two randomly sampled generations from different models. We then ask the annotator which generations they prefer.

%Two of the authors\footnote{anonymized background}, with the help of the mayor of one of the 20th largest cities of Switzerland, annotated 40 comments 

\begin{figure}[h!]
    \centering
    \includegraphics[width=0.9\linewidth]{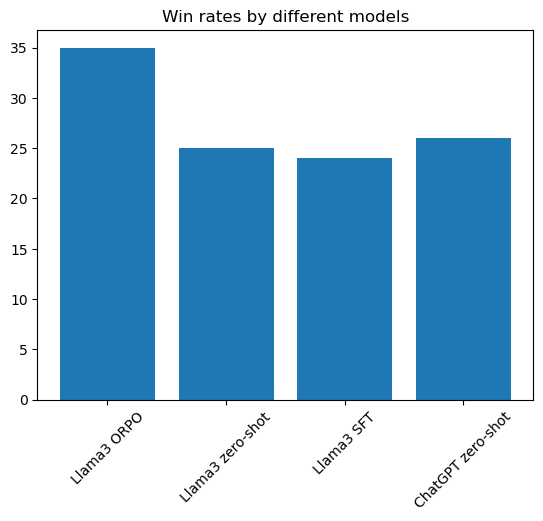}
    \caption{Win rates by different models.}
    \label{fig:seq_lengths}
\end{figure}

Due to the high costs of manual evaluation, we only select four models for human validation. Overall, the generations produced by the ORPO-aligned models are preferred in around 60\% of the cases, whereas generations from the other models have similar win rates. The author team, with the help of the mayor of a Swiss city, manually annotated 40 comments in a deliberative setting. All data points during that annotation round were discussed at length. Before seeing the model generations, the team discussed what an optimal generation for a question given the party would look like. We then discussed which of the two generations is closer to that. We treat this set as the gold standard. Inter annotator agreement of the annotator and this gold standard is 0.55 (Cohen's kappa), indicating moderate agreement. If we discard evaluations where the annotator or the team settles for a tie, this agreement rises to 0.84, indicating almost perfect agreement. Detailed instructions, annotator demographics, and further robustness validations can be found in Appendix \ref{sec:annotation_guidelines}.

In sum, all evaluations suggest that aligning LLMs with ORPO seems to work best to generate diverse political viewpoints. But what are the reasons why annotators prefer the generations of the ORPO-aligned model?

We analyze all human evaluations where the ORPO-aligned model is involved and the human annotation is not a tie. This error analysis reveals that we can categorize these data points as follows: The losing model's generation is (a) inaccurate (n=35), (b) less nuanced (n=27), or (c) otherwise inferior (n=6).\footnote{Category (c) contains low readability, bad grammar, or generation in the wrong language.} We find that the ORPO-aligned model's response was often preferred because it is more accurate (60\% winrate in these cases). At the same time, the generation of the ORPO-aligned model is often slightly less nuanced (44\% winrate in cases where the decision was explained by nuance). For (c), ORPO-aligned models are neither doing worse or better than other models (50\%).

\section{Discussion and Conclusion}

If explicitly asked to produce political viewpoints from a certain party perspective, LLM-generated text should accurately reflect the viewpoints of the party. We show that current LLMs fail to do so in a zero-shot setting in the Swiss context.

Combining alignment and conditional generation substantially improves such generations, as we have shown qualitatively and quantitatively throughout this work. Supervised fine-tuning of Llama models and few-shot experiments result in similar performance, both beating the zero-shot setting. ORPO-aligned models work best for this type of conditional generation, and we confirm this finding with a diverse set of results.

Such aligned models have practical use cases beyond accurately presenting specific party preferences. We can also use such models to generate a balanced overview of viewpoints toward a specific issue. We show a simple algorithm to do so in Figure \ref{fig:pseudocode}.

\begin{figure}[h!]
    \small
    \centering
    \begin{algorithmic}[1]
        \State \textbf{Initialize} answers = Ø
        \For{\textbf{each} party \textbf{in} parties}
            \State answer = generate\_answer(LLM, party)
            \State answers.append(answer)
        \EndFor
        \State synthesize\_answers(gpt4, answers)
    \end{algorithmic}
    \caption{Pseudocode for generating and synthesizing answers.}
    \label{fig:pseudocode}
\end{figure}

We first generate a stance to a policy issue for all parties, and then let GPT-4o provide a summary of these stances. For illustration, we run this procedure for the issue \textit{Should the state do more to promote equal educational opportunities?} Figure \ref{fig:example_synthesized} shows the overview synthesized from the ORPO-aligned Llama 3 model is more balanced and accurate than the one synthesized from the responses obtained in a zero-shot setting.

Generating text that contains political views has major implications. LLMs have the potential to shift attitudes and behavior. If they do that in the political domain, this might influence elections, one of the most important decision-making processes in democracies \citep{berger_et_al_2008}.
Our recommendations are two-fold: First, further research is necessary to explore the promises and pitfalls of LLMs delivering political information or offering explicit voting advice. Second, there is a need for both societal and scientific debates on the role of LLMs and AI in democratic processes. %We elaborated on these points in the Ethics Section.

\citet{hartmann2023political} asked what if ChatGPT exhibited a political ideology that may pervade its synthetic responses and subtly influence its unsuspecting users? Our work speaks to this question. 

We see a number of strategies going forward: (1) LLMs would always refuse to answer anything related to shaping political beliefs and take sincere political impartiality as an alignment goal. (2) LLMs would always produce broad overviews of political issues (as produced by our methods in Figure \ref{fig:example_synthesized}). Our work might facilitate creating appropriate datasets for this. (3) LLMs would explicitly produce text that is aligned with a certain political leaning. In this case, however, the provider of an LLM would be fully transparent about this, and/or the user should be able to fully control what ideology LLM-generated text should be aligned with. Aligning models explicitly with party preferences using conditional generation, as presented in this work, is one way toward such LLMs.

\newpage

\section*{Limitations}\label{sec:limitations}

We acknowledge several limitations and outline a range of possibilities for future work.

\paragraph{Choice of models and Alignment algorithms.} We have mainly experimented with Llama 3 models and ChatGPT 3.5 zero-shot. There are, by now, other capable open-source models (Mistral, Mixtral, Llama 2) or model sizes (70B) that we could have fine-tuned with the method proposed in this work. Also, there exists a range of different alignment algorithms (DPO, RLHF), which we have experimented with, but the resulting models did not pass initial vibes tests. We plan to investigate all of this more thoroughly in future work.
\paragraph{Choice of metadata for conditional generation.} In preliminary experiments, we experimented with the \citep{vamvas2020xstance} dataset and generated comments based on stance (pro/contra) and not party affiliation. Eyeballing these results indicates that ORPO-aligned models in this setting also return more diverse answers than zero-shot models, and ORPO-aligned model generations seem more creative than SFT models. We take this as evidence for the robustness of conditional alignment, but we have not exhaustively evaluated this. Next, we think there are exciting opportunities in alignment with more metadata, such as canton, age, gender, and any other attribute potentially influencing political viewpoints. We tried this in preliminary experiments with Mistral models and DPO. However, this didn't work. We plan to revisit this with Llama 3 and ORPO.
\paragraph{Data availability.} All data used in this study and the replication package are publicly available on github: \href{https://github.com/dominiksinsaarland/swiss_alignment}{dominiksinsaarland/swiss\_alignment}.
\paragraph{Further data.} We believe it should be possible, in principle, to find more diverse data sources with party affiliation (e.g., newspaper or TV interviews, party website content). It should be possible to collect such, and this would make for a dataset including more diverse political questions, which might lead to more creative models. This approach can be used across countries and parties, and thus allow for replication studies in contexts not related to smartvote data or Switzerland.
\newpage

\section*{Ethics Statement}\label{sec:ethics_statement}
We also acknowledge ethical implications of our work.

\paragraph{Contested topic.} We believe the combination of LLMs and democracy is a very delicate topic, and throughout the manuscript tried to do justice to such challenging circumstances. On one hand, we all exhibit political biases, which we tried to remove from the paper as good as possible, but we also acknowledge that we probably have not written a completeley impartial piece. The same holds for LLMs. Another goal of this paper is to increase the awareness about these points.

\paragraph{Intended use case.} We believe it is important that LLMs, its users, developers testers and other stakeholders are aware of political bias in machine-generated text. In this paper, we do not argue for creating chatbots which act as echochambers and/or reinforce existing biases present in such models -- or change political views or actions of users. We want to argue for the opposite, that LLMs should exactly not do that. We tried to do justice to this goal.

\paragraph{Biases in LLMs.} Political bias is one sort of bias present in LLMs. There are others \citep[see e.g.,][]{abid2021persistent,lucy2021gender}, which are not addressed in this work. Our resulting models may potentially perpetuate these biases.

\paragraph{Accuracy, hallucinations, and outdated information.} Our aligned models, as well as ChatGPT, are not 100\% accurate in producing political information: They produce hallucinations or other potentially harmful text, hence we do not advocate to use them in a commercial context, but propose a method to potentially mitigate political biases in LLMs. Further, we align our models on smartvote comments from 2015 - 2023. Parties might change their stance in the meantime. It remains an open research question how to incorporate such changes in stances.

\paragraph{Non-constitutional parties.} We included viewpoints of political parties that operate within the limits of the constitution. Whether LLMs should reproduce the content of extremist parties without disclaimers is not within the scope of our research.

\bibliography{custom}

\begin{thebibliography}{34}
\providecommand{\natexlab}[1]{#1}

\bibitem[{Abid et~al.(2021)Abid, Farooqi, and Zou}]{abid2021persistent}
Abubakar Abid, Maheen Farooqi, and James Zou. 2021.
\newblock \href {https://dl.acm.org/doi/abs/10.1145/3461702.3462624} {{Persistent Anti-Muslim Bias in Large Language Models}}.
\newblock In \emph{Proceedings of the 2021 AAAI/ACM Conference on AI, Ethics, and Society}, pages 298--306.

\bibitem[{AI@Meta(2024)}]{llama3modelcard}
AI@Meta. 2024.
\newblock \href {https://github.com/meta-llama/llama3/blob/main/MODEL_CARD.md} {{Llama 3 Model Card}}.

\bibitem[{Amplayo and Lapata(2021)}]{amplayo-lapata-2021-informative}
Reinald~Kim Amplayo and Mirella Lapata. 2021.
\newblock \href {https://doi.org/10.18653/v1/2021.eacl-main.229} {{Informative and Controllable Opinion Summarization}}.
\newblock In \emph{Proceedings of the 16th Conference of the European Chapter of the Association for Computational Linguistics: Main Volume}, pages 2662--2672, Online. Association for Computational Linguistics.

\bibitem[{Bakker et~al.(2022)Bakker, Chadwick, Sheahan, Tessler, Campbell-Gillingham, Balaguer, McAleese, Glaese, Aslanides, Botvinick, and Summerfield}]{consent_statements}
Michiel Bakker, Martin Chadwick, Hannah Sheahan, Michael Tessler, Lucy Campbell-Gillingham, Jan Balaguer, Nat McAleese, Amelia Glaese, John Aslanides, Matt Botvinick, and Christopher Summerfield. 2022.
\newblock \href {https://proceedings.neurips.cc/paper_files/paper/2022/file/f978c8f3b5f399cae464e85f72e28503-Paper-Conference.pdf} {{Fine-tuning Language Models to Find Agreement among Humans with Diverse Preferences}}.
\newblock In \emph{Advances in Neural Information Processing Systems}, volume~35, pages 38176--38189. Curran Associates, Inc.

\bibitem[{Bang et~al.(2024)Bang, Chen, Lee, and Fung}]{bang-etal-2024-measuring}
Yejin Bang, Delong Chen, Nayeon Lee, and Pascale Fung. 2024.
\newblock \href {https://aclanthology.org/2024.acl-long.600} {{Measuring Political Bias in Large Language Models: What Is Said and How It Is Said}}.
\newblock In \emph{Proceedings of the 62nd Annual Meeting of the Association for Computational Linguistics (Volume 1: Long Papers)}, pages 11142--11159, Bangkok, Thailand. Association for Computational Linguistics.

\bibitem[{Berger et~al.(2008)Berger, Meredith, and Wheeler}]{berger_et_al_2008}
Jonah Berger, Marc Meredith, and S.~Christian Wheeler. 2008.
\newblock \href {https://doi.org/10.1073/pnas.0711988105} {{Contextual Priming: Where People Vote Affects how they Vote}}.
\newblock \emph{Proceedings of the National Academy of Sciences}, 105(26):8846--8849.

\bibitem[{Conneau et~al.(2020)Conneau, Khandelwal, Goyal, Chaudhary, Wenzek, Guzm{\'a}n, Grave, Ott, Zettlemoyer, and Stoyanov}]{conneau-etal-2020-unsupervised}
Alexis Conneau, Kartikay Khandelwal, Naman Goyal, Vishrav Chaudhary, Guillaume Wenzek, Francisco Guzm{\'a}n, Edouard Grave, Myle Ott, Luke Zettlemoyer, and Veselin Stoyanov. 2020.
\newblock \href {https://doi.org/10.18653/v1/2020.acl-main.747} {{Unsupervised Cross-lingual Representation Learning at Scale}}.
\newblock In \emph{Proceedings of the 58th Annual Meeting of the Association for Computational Linguistics}, pages 8440--8451, Online. Association for Computational Linguistics.

\bibitem[{Devlin et~al.(2019)Devlin, Chang, Lee, and Toutanova}]{devlin-etal-2019-bert}
Jacob Devlin, Ming-Wei Chang, Kenton Lee, and Kristina Toutanova. 2019.
\newblock \href {https://doi.org/10.18653/v1/N19-1423} {{BERT: Pre-training of Deep Bidirectional Transformers for Language Understanding}}.
\newblock In \emph{Proceedings of the 2019 Conference of the North {A}merican Chapter of the Association for Computational Linguistics: Human Language Technologies, Volume 1 (Long and Short Papers)}, pages 4171--4186, Minneapolis, Minnesota. Association for Computational Linguistics.

\bibitem[{Feng et~al.(2023)Feng, Park, Liu, and Tsvetkov}]{feng-etal-2023-pretraining}
Shangbin Feng, Chan~Young Park, Yuhan Liu, and Yulia Tsvetkov. 2023.
\newblock \href {https://doi.org/10.18653/v1/2023.acl-long.656} {{From Pretraining Data to Language Models to Downstream Tasks: Tracking the Trails of Political Biases Leading to Unfair NLP Models}}.
\newblock In \emph{Proceedings of the 61st Annual Meeting of the Association for Computational Linguistics (Volume 1: Long Papers)}, pages 11737--11762, Toronto, Canada. Association for Computational Linguistics.

\bibitem[{Feng et~al.(2024)Feng, Sorensen, Liu, Fisher, Park, Choi, and Tsvetkov}]{feng2024modularpluralismpluralisticalignment}
Shangbin Feng, Taylor Sorensen, Yuhan Liu, Jillian Fisher, Chan~Young Park, Yejin Choi, and Yulia Tsvetkov. 2024.
\newblock \href {https://arxiv.org/abs/2406.15951} {{Modular Pluralism: Pluralistic Alignment via Multi-LLM Collaboration}}.
\newblock \emph{Preprint}, arXiv:2406.15951.

\bibitem[{Hartmann et~al.(2023)Hartmann, Schwenzow, and Witte}]{hartmann2023political}
Jochen Hartmann, Jasper Schwenzow, and Maximilian Witte. 2023.
\newblock \href {https://arxiv.org/abs/2301.01768} {{The Political Ideology of Conversational AI: Converging Evidence on ChatGPT's Pro-environmental, Left-libertarian Orientation}}.
\newblock \emph{Preprint}, arXiv:2301.01768.

\bibitem[{Hong et~al.(2024)Hong, Lee, and Thorne}]{hong2024orpo}
Jiwoo Hong, Noah Lee, and James Thorne. 2024.
\newblock \href {https://arxiv.org/abs/2403.07691} {{ORPO: Monolithic Preference Optimization without Reference Model}}.
\newblock \emph{Preprint}, arXiv:2403.07691.

\bibitem[{Hu et~al.(2022)Hu, Shen, Wallis, Allen-Zhu, Li, Wang, Wang, and Chen}]{hu2022lora}
Edward~J Hu, Yelong Shen, Phillip Wallis, Zeyuan Allen-Zhu, Yuanzhi Li, Shean Wang, Lu~Wang, and Weizhu Chen. 2022.
\newblock \href {https://openreview.net/forum?id=nZeVKeeFYf9} {{LORA: Low-Rank Adaptation of Large Language Models}}.
\newblock In \emph{International Conference on Learning Representations}.

\bibitem[{Jakesch et~al.(2023)Jakesch, Bhat, Buschek, Zalmanson, and Naaman}]{Jakesch2023}
Maurice Jakesch, Advait Bhat, Daniel Buschek, Lior Zalmanson, and Mor Naaman. 2023.
\newblock \href {https://doi.org/10.1145/3544548.3581196} {{Co-Writing with Opinionated Language Models Affects Users’ Views}}.
\newblock In \emph{Proceedings of the 2023 CHI Conference on Human Factors in Computing Systems}, CHI '23, New York, NY, USA. Association for Computing Machinery.

\bibitem[{Jiang et~al.(2022)Jiang, Beeferman, Roy, and Roy}]{jiang-etal-2022-communitylm}
Hang Jiang, Doug Beeferman, Brandon Roy, and Deb Roy. 2022.
\newblock \href {https://aclanthology.org/2022.coling-1.593} {{CommunityLM: Probing Partisan Worldviews from Language Models}}.
\newblock In \emph{Proceedings of the 29th International Conference on Computational Linguistics}, pages 6818--6826, Gyeongju, Republic of Korea. International Committee on Computational Linguistics.

\bibitem[{Lucy and Bamman(2021)}]{lucy2021gender}
Li~Lucy and David Bamman. 2021.
\newblock \href {https://aclanthology.org/2021.nuse-1.5/} {{Gender and Representation Bias in {GPT}-3 Generated Stories}}.
\newblock In \emph{Proceedings of the Third Workshop on Narrative Understanding}, pages 48--55.

\bibitem[{Mehdi(2023)}]{mehdi2023}
Yusuf Mehdi. 2023.
\newblock {Reinventing Search with a New AI-powered Microsoft Bing and Edge, Your Copilot for the Web}.
\newblock Accessed: 2024-06-08.

\bibitem[{Motoki et~al.(2024)Motoki, Pinho~Neto, and Rodrigues}]{Motoki2024}
Fabio Motoki, Valdemar Pinho~Neto, and Victor Rodrigues. 2024.
\newblock \href {https://doi.org/10.1007/s11127-023-01097-2} {{More Human Than Human: Measuring ChatGPT Political Bias}}.
\newblock \emph{Public Choice}, 198(1):3--23.

\bibitem[{OpenAI et~al.(2024)OpenAI, Achiam, Adler, Agarwal, Ahmad, and et~al.}]{openai2024gpt4}
OpenAI, Josh Achiam, Steven Adler, Sandhini Agarwal, Lama Ahmad, and Ilge~Akkaya et~al. 2024.
\newblock \href {https://arxiv.org/abs/2303.08774} {{GPT-4 Technical Report}}.
\newblock \emph{Preprint}, arXiv:2303.08774.

\bibitem[{Pillutla et~al.(2021)Pillutla, Swayamdipta, Zellers, Thickstun, Welleck, Choi, and Harchaoui}]{pillutla-etal:mauve:neurips2021}
Krishna Pillutla, Swabha Swayamdipta, Rowan Zellers, John Thickstun, Sean Welleck, Yejin Choi, and Zaid Harchaoui. 2021.
\newblock {MAUVE: Measuring the Gap Between Neural Text and Human Text using Divergence Frontiers}.
\newblock In \emph{NeurIPS}.

\bibitem[{Rafailov et~al.(2023)Rafailov, Sharma, Mitchell, Ermon, Manning, and Finn}]{rafailov2023direct}
Rafael Rafailov, Archit Sharma, Eric Mitchell, Stefano Ermon, Christopher~D. Manning, and Chelsea Finn. 2023.
\newblock \href {https://arxiv.org/abs/2305.18290} {{Direct Preference Optimization: Your Language Model is Secretly a Reward Model}}.
\newblock \emph{Preprint}, arXiv:2305.18290.

\bibitem[{Rozado(2023)}]{rozado2023political}
David Rozado. 2023.
\newblock {The Political Biases of ChatGPT}.
\newblock \emph{Social Sciences}, 12(3):148.

\bibitem[{Rutinowski et~al.(2024)Rutinowski, Franke, Endendyk, Dormuth, Roidl, and Pauly}]{Rutinowski2024}
Jérôme Rutinowski, Sven Franke, Jan Endendyk, Ina Dormuth, Moritz Roidl, and Markus Pauly. 2024.
\newblock \href {https://doi.org/10.1155/2024/7115633} {{The Self-Perception and Political Biases of ChatGPT}}.
\newblock \emph{Human Behavior and Emerging Technologies}, 2024(1):7115633.

\bibitem[{Sharma et~al.(2024)Sharma, Liao, and Xiao}]{sharma2024generativeechochambereffects}
Nikhil Sharma, Q.~Vera Liao, and Ziang Xiao. 2024.
\newblock \href {https://arxiv.org/abs/2402.05880} {{Generative Echo Chamber? Effects of LLM-Powered Search Systems on Diverse Information Seeking}}.
\newblock \emph{Preprint}, arXiv:2402.05880.

\bibitem[{Stieger et~al.(2021)Stieger, Flückiger, Rüegger, Kowatsch, Roberts, and Allemand}]{stieger2021}
Mirjam Stieger, Christoph Flückiger, Dominik Rüegger, Tobias Kowatsch, Brent~W. Roberts, and Mathias Allemand. 2021.
\newblock \href {https://doi.org/10.1073/pnas.2017548118} {{Changing Personality Traits with the help of a Digital Personality Change Intervention}}.
\newblock \emph{Proceedings of the National Academy of Sciences}, 118(8):e2017548118.

\bibitem[{Stiennon et~al.(2020)Stiennon, Ouyang, Wu, Ziegler, Lowe, Voss, Radford, Amodei, and Christiano}]{stiennon2022learning}
Nisan Stiennon, Long Ouyang, Jeff Wu, Daniel~M. Ziegler, Ryan Lowe, Chelsea Voss, Alec Radford, Dario Amodei, and Paul Christiano. 2020.
\newblock {Learning to Summarize from Human Feedback}.
\newblock In \emph{Proceedings of the 34th International Conference on Neural Information Processing Systems}, NIPS '20, Red Hook, NY, USA. Curran Associates Inc.

\bibitem[{Suhara et~al.(2020)Suhara, Wang, Angelidis, and Tan}]{suhara-etal-2020-opiniondigest}
Yoshihiko Suhara, Xiaolan Wang, Stefanos Angelidis, and Wang-Chiew Tan. 2020.
\newblock \href {https://doi.org/10.18653/v1/2020.acl-main.513} {{OpinionDigest: A Simple Framework for Opinion Summarization}}.
\newblock In \emph{Proceedings of the 58th Annual Meeting of the Association for Computational Linguistics}, pages 5789--5798, Online. Association for Computational Linguistics.

\bibitem[{Taori et~al.(2023)Taori, Gulrajani, Zhang, Dubois, Li, Guestrin, Liang, and Hashimoto}]{taori2023alpaca}
Rohan Taori, Ishaan Gulrajani, Tianyi Zhang, Yann Dubois, Xuechen Li, Carlos Guestrin, Percy Liang, and Tatsunori~B Hashimoto. 2023.
\newblock {Alpaca: A Strong, Replicable Instruction-following Model}.
\newblock \emph{Stanford Center for Research on Foundation Models. https://crfm. stanford. edu/2023/03/13/alpaca. html}, 3(6):7.

\bibitem[{Thurman and Gasser(2009)}]{thurman2009smartvote}
James Thurman and Urs Gasser. 2009.
\newblock {Three Case Studies from Switzerland: Smartvote}.
\newblock Technical report, Berkman Center Research Publications.

\bibitem[{Tunstall et~al.(2023)Tunstall, Beeching, Lambert, Rajani, Rasul, Belkada, Huang, von Werra, Fourrier, Habib, Sarrazin, Sanseviero, Rush, and Wolf}]{tunstall2023zephyr}
Lewis Tunstall, Edward Beeching, Nathan Lambert, Nazneen Rajani, Kashif Rasul, Younes Belkada, Shengyi Huang, Leandro von Werra, Clémentine Fourrier, Nathan Habib, Nathan Sarrazin, Omar Sanseviero, Alexander~M. Rush, and Thomas Wolf. 2023.
\newblock \href {https://arxiv.org/abs/2310.16944} {{Zephyr: Direct Distillation of LM Alignment}}.
\newblock \emph{Preprint}, arXiv:2310.16944.

\bibitem[{Vamvas and Sennrich(2020)}]{vamvas2020xstance}
Jannis Vamvas and Rico Sennrich. 2020.
\newblock \href {https://arxiv.org/abs/2003.08385} {{X-Stance: A Multilingual Multi-Target Dataset for Stance Detection}}.
\newblock \emph{Preprint}, arXiv:2003.08385.

\bibitem[{von Werra et~al.(2020)von Werra, Belkada, Tunstall, Beeching, Thrush, Lambert, and Huang}]{vonwerra2022trl}
Leandro von Werra, Younes Belkada, Lewis Tunstall, Edward Beeching, Tristan Thrush, Nathan Lambert, and Shengyi Huang. 2020.
\newblock {TRL: Transformer Reinforcement Learning}.
\newblock \url{https://github.com/huggingface/trl}.

\bibitem[{Zellers et~al.(2019)Zellers, Holtzman, Rashkin, Bisk, Farhadi, Roesner, and Choi}]{NEURIPS2019_3e9f0fc9}
Rowan Zellers, Ari Holtzman, Hannah Rashkin, Yonatan Bisk, Ali Farhadi, Franziska Roesner, and Yejin Choi. 2019.
\newblock \href {https://proceedings.neurips.cc/paper_files/paper/2019/file/3e9f0fc9b2f89e043bc6233994dfcf76-Paper.pdf} {{Defending Against Neural Fake News}}.
\newblock In \emph{Advances in Neural Information Processing Systems}, volume~32. Curran Associates, Inc.

\bibitem[{Zhou et~al.(2023)Zhou, Jiang, Wilcox, Cotterell, and Sachan}]{zhou2023controlled}
Wangchunshu Zhou, Yuchen~Eleanor Jiang, Ethan Wilcox, Ryan Cotterell, and Mrinmaya Sachan. 2023.
\newblock \href {https://arxiv.org/abs/2304.14293} {{Controlled Text Generation with Natural Language Instructions}}.
\newblock \emph{Preprint}, arXiv:2304.14293.

\end{thebibliography}

\newpage
\clearpage
\appendix

\section{ChatGPT and smartvote}\label{sec:political_bias_chatgpt}

We have taken the smartVote survey with ChatGPT for the national elections in 2023 in Switzerland. We have taken the short survey (30 questions) using a temperature of 0 and the following system prompt: \textit{You are a helpful research assistant in Switzerland. You are given a few questions which you need to answer. You MUST reply with only one of {"yes", "rather yes", "rather no", "no"}.} The output categories are the options users can enter while taking the smartvote survey. The user prompt equals to the actual smart vote questions, e.g., "Should the state do more to promote equal educational opportunities"?

\begin{figure}
    \centering
    \includegraphics[width=\linewidth]{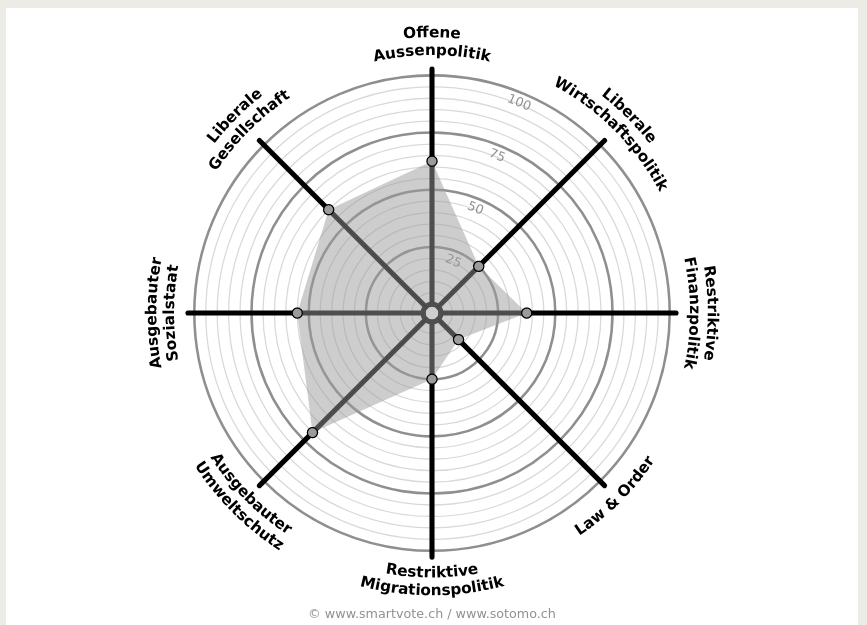}
    \caption{Overview of political preferences of ChatGPT in Switzerland for the 2023 national elections of parliament (source: smartvote.ch)}
    \includegraphics[width=\linewidth]{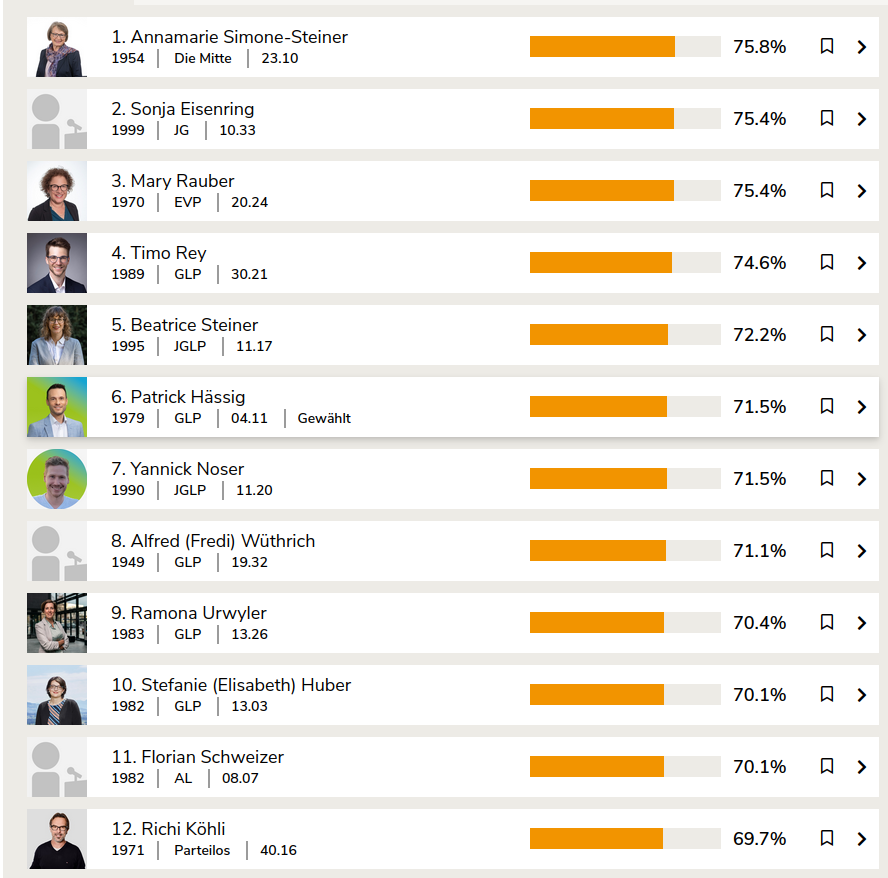}
    Swiss candidates have the highest overlap in political preferences regarding ChatGPT (source: smartvote.ch). 
    \label{fig:smartvote_output}
\end{figure}

In Figure \ref{fig:smartvote_output}, we show the output a user receives after taking the smartvote survey: In the top panel, we show political preferences across different dimensions. The categories correspond to: Liberale Gesellschaft = Liberal society; Offene Aussenpolitik = Open foreign policy; Liberale Wirtschaftspolitik = Liberal economic policy; Restriktive Finanzpolitik =Restrictive financial policy; Restriktive Migrationspolitk = Restrictive migration policy; Ausgebauter Umweltschutz = Expanded environmental protection; Ausgebauter Sozialstaat = Expanded welfare state.

In the bottom panel, we show the candidates who were identified as having the highest political overlap with ChatGPT. 7 out of 12 (58\%) of the most aligned candidates would be from the (Young) Liberal Party of Switzerland (GLP or JGLP).

\section{Dataset Statistics}
\label{sec:dataset_stats}

In Table \ref{tab:dataset_examples}, we show 5 randomly sampled comments (and their English translation) from our dataset.

In Table \ref{tab:dataset_stats}, we show dataset statistics, the number of examples in each split, the number of political issues and the share of different languages in the different splits of the dataset. In Figure \ref{fig:seq_lengths}, we show a histogram of the sequence lengths in the dataset (across all splits). We excluded comments shorter than five words.

We have access to smartvote data for the national parliament elections in 2015, 2019, and 2023. We split the data into a training set, a development, and a test set. Both the development and test set consist of 10\% of the political issues from the 2023 election that were not present in the 2015 or 2019 survey.

\begin{table}[h!]
    \centering
    \tiny
    \begin{tabular}{p{2.3cm} p{2cm} p{2.3cm}}
    issue & comment (English) & comment (original)  \\ \hline
Are you in favor of amending the social welfare guidelines to reduce benefits for large families and young adults? & I find it important to look at the individual persons/families concerned. & Auch hier finde ich die individuelle Betrachtung der betreffenden Personen/Familien wichtig.  \\
Should the state provide more funds for health insurance premium reductions? & Yes, the state must invest massively more in combating rising poverty. & Ja, der Staat muss massiv mehr in die Bekämpfung der steigenden Armut investieren. \\
Should incentives and target agreements rather than bans and restrictions be used exclusively to achieve the climate targets? & To guide certain behaviors, however, the time has come for prohibitions and restrictions. & Pour guider certains comportement, l'heure est quand même aux interdictions et restrictions. \\
Should the differences between financially strong and weak cantons be reduced more through financial equalization? & Wealthy cantons have benefited greatly from corporate tax cuts in recent years. & Les cantons riches ont ces dernières années largement profité des réductions de l'imposition des entreprises. \\
The financially strong cantons would like to significantly reduce their contributions to the financially weak cantons as part of the financial equalization (NFA). Do you support this request? & Long-term abuse of solidarity is counterproductive. & Un abuso della solidarietà a lungo termine è contraproducente. \\ \hline
    \end{tabular}
    \caption{Example comments from the dataset (automatically translated with deepl, and manually checked whether the translation is accurate).}
    \label{tab:dataset_examples}
\end{table}

\begin{table}[h!]
    \centering
    \tiny
    \begin{tabular}{l c c c c c}
    split &  \# examples & \# political issues & German (\%) & French (\%) & Italian (\%) \\ \hline 
    train & 92,986 & 203 & 75.5 & 22.2 & 2.2 \\ 
    dev & 4262 & 7 & 76.8 & 21.0 & 2.3 \\
    test & 5488 & 7 & 78.4 & 19.3 & 2.2 \\ \hline
    \end{tabular}
    \caption{Dataset Statistics}
    \label{tab:dataset_stats}
\end{table}

\begin{figure}
    \centering
    \includegraphics[width=\linewidth]{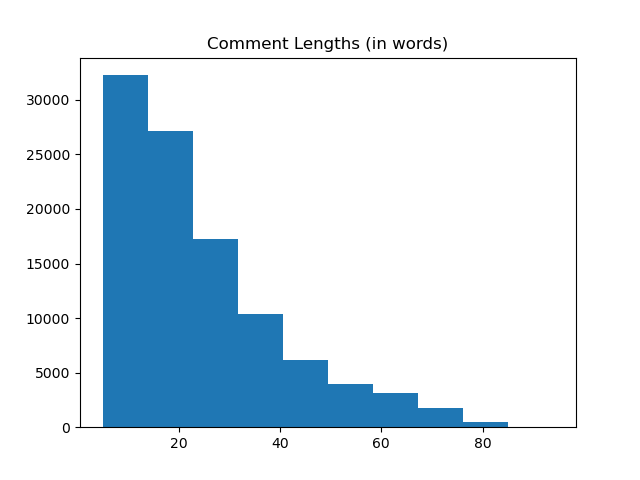}
    \caption{Sequence lengths of Smartvote comments.}
    \label{fig:seq_lengths}
\end{figure}

We show the 10 most often occuring parties and their associated number of comments in Table \ref{tab:party_frequencies}.

\begin{table}[]
    \centering
    \begin{tabular}{l c}
party & \# comments \\ \hline
FDP &  15589 \\
GLP &  11341 \\
GRÜNE &  8992 \\
SP &  8880 \\
EVP &  7734 \\
SVP &  6780 \\
DIE MITTE &  6274 \\
CVP &  4756 \\
EDU &  3940 \\
JG &  3595 \\ \hline
    \end{tabular}
    \caption{Caption}
    \label{tab:party_frequencies}
\end{table}
\section{Annotation Guidelines}\label{sec:annotation_guidelines}

We recruited an annotator from Switzerland with a university degree in political science and a strong self-declared interest in Swiss politics. The annotator read a random sample of 200 messages. The annotator was instructed as follows:

\vskip 2mm

\noindent \textit{Here is some information on the project: Large Language Models (LLMs, such as ChatGPT) often have biases. For a research project, we have fine-tuned an open-source LLM to make it more representative of the political values of different Swiss people. We used Smartvote data for this alignment. Now, an important question is whether our fine-tuned model is better than other models in terms of how accurately it presents the viewpoints of Swiss politicians. This is where your contribution is important. In the linked file, you will see:}

\begin{itemize}
\item \textit{A prompt}
\item \textit{Two responses from LLMs (Candidate A and Candidate B)}
\item \textit{Reference comments}
\item \textit{Two columns that you will fill in (see instructions below)}
\end{itemize}

\vskip 2mm
\textit{\textbf{First column to fill in: your preference.} Please ask yourself if Candidate A or B is better. Here, "better" means that the respective Candidate most closely reflects the reference comments and aligns with your knowledge of that party's stance on the respective policy question. You can also leverage your knowledge of whether a given position would correspond to that party's mainstream position.}

\begin{itemize}
    \item \textit{1 = both A and B are good}
    \item \textit{0: neither A nor B is good}
    \item \textit{A = only A is good, or A is significantly better than B
    \item B = only B is good, or B is significantly better than A}
\end{itemize}

\textit{Some illustrations:
If both are OK/good, but one is better, you will enter your favorite.
If both are good and it is impossible to choose between A and B, you enter "1".
If none are good, enter "0".}

\vskip 2mm
\textit{\textbf{Second column to fill in: your familiarity.} Here you would indicate how familiar you are with the party's position on the issue.}

\begin{itemize}
\item \textit{1 = wild guess/need to read more about it}
\item \textit{2 = have a clue, but need to double-check}
\item \textit{3 = reasonable guess}
\item \textit{4 = quite sure about the party's stance on issue Q, I would be surprised if the stance is different from my intuition}
\item \textit{5 = 100\% sure about the party's stance}
\end{itemize}

\subsection{Further Robustness Checks in Human Annotations}

We drop annotation datapoints where the annotator indicated that the familiarity with a political issue and the party stance would be less than a reasonable guess. We find that the inter annotator agreement (measured in Cohen's kappa) for these annotated datapoints and our gold standard is 0.34, and thus substantially lower than the presented main results. Given these reasons, we don't think these datapoints are a reliable source for measuring preferences.

\begin{figure}[h!]
    \centering
    \includegraphics[width=0.75\linewidth]{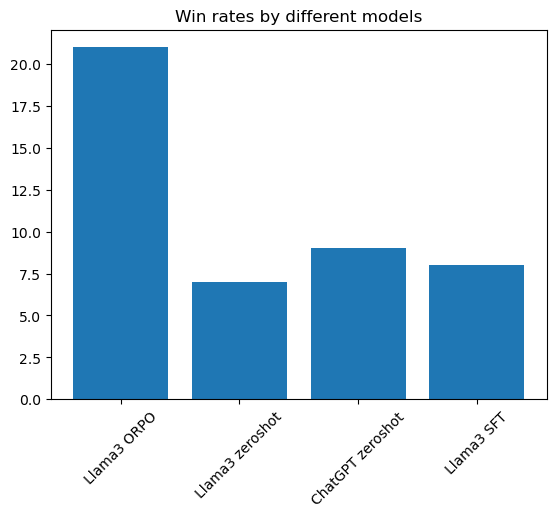}
    \caption{Win rates by different models where annotator is confident. }
    \label{fig:confident_annotations}
\end{figure}

\begin{figure}[h!]
    \centering
    \includegraphics[width=0.75\linewidth]{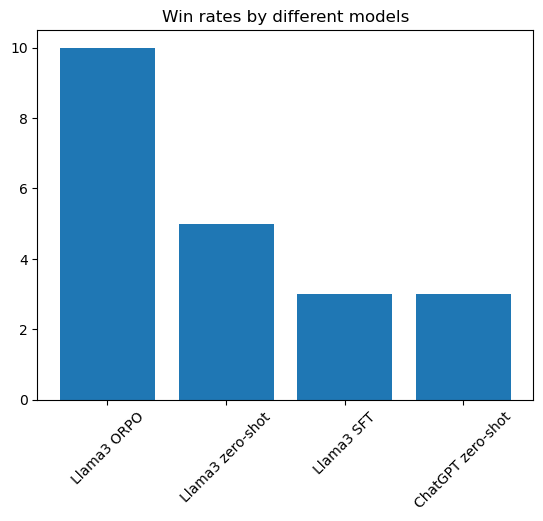}
    \caption{Win rates by different models (gold standard).}
    \label{fig:gold_standard_annotations}
\end{figure}

If we only consider data points where the annotator is fairly confident ("quite sure about the party's stance on issue Q" or "100\% sure about the party's stance"), the inter annotator agreement raises to 0.61, indicating moderate agreement. In this setting, annotator preference for ORPO-aligned generations is even more pronounced. We show these win rates in Figure \ref{fig:confident_annotations}. Lastly, we show overall win rates by models in our gold standard in Figure \ref{fig:gold_standard_annotations}.

\section{Generating Overviews over Political Issues}

\begin{figure}[h!]
    \vspace{-1mm}\hspace*{-2.7mm}
    \centering
    \tiny
    \begin{tabular}{p{7.7cm}} 
\textbf{Generated overview based on ChatGPT replies:} In Switzerland, there is a broad consensus among major political parties that the state should take more action to promote equal educational opportunities. The common rationale is to foster equality and reduce social disparities. A widely supported measure is the introduction of tutoring vouchers for students from low-income families, which is seen as an effective tool to level the educational playing field and provide all children with equal starting opportunities. This approach is endorsed across the political spectrum, reflecting a unified commitment to enhancing educational equity. \\ 
\textbf{Generated overview based on ORPO-aligned Llama 3 replies:} The question of whether the state should do more to promote equal educational opportunities is a significant topic in Swiss politics, with varied perspectives across the political spectrum. Some parties emphasize the importance of education as a fundamental right and a cornerstone of societal equity, advocating for enhanced state involvement to ensure equal opportunities from early education onwards. Others argue that the responsibility for education lies primarily with the cantons, suggesting that local governance is better suited to address educational needs. Some views focus on the importance of supporting both high and low-performing students, while some parties believe that the current educational opportunities in Switzerland are already sufficient and that the focus should instead be on integration. Overall, the debate reflects a balance between the need for national standards and local autonomy, as well as differing opinions on the best methods to achieve educational equity.
    \end{tabular}
    \vspace{-2.5mm}
    \caption{Prompt for conditional generation. Varying attributes are party, language and political issue.}
    \label{fig:example_synthesized}
\end{figure}

\end{document}